%% file: main.tex
\documentclass[letterpaper, 10 pt, conference]{ieeeconf} 
\IEEEoverridecommandlockouts
\input{preamble.tex}
\title{\LARGE \bf Versatile Locomotion Skills for Hexapod Robots}
\author{Tomson Qu$^{1}$, Dichen Li$^{1}$, Avideh Zakhor$^{1}$, Wenhao Yu$^{2}$, Tingnan Zhang$^{2}$
\thanks{$^{1}$ EECS Department, University of California, Berkeley}
\thanks{$^{2}$ Google DeepMind}}

\begin{document}
\maketitle
\thispagestyle{empty}
\pagestyle{empty}
\begin{abstract}
\input{abstract}

\end{abstract}
\input{section-1-introduction}

\input{section-2-related-work}
\input{section-4-methodology}

\input{section-5-experiments}
\input{section-7-limitations-and-future-work}
\input{section-8-conclusion}

\bibliographystyle{IEEEtran}
\bibliography{references}

\end{document}

%% file: preamble.tex
\usepackage[utf8]{inputenc}
\usepackage{graphicx}
\usepackage{multicol}
\usepackage{multirow}
\usepackage{array}
\usepackage{footmisc}
\usepackage{amssymb}
\usepackage{amsmath}
\usepackage{amsfonts}
\usepackage{algorithm}
\usepackage{algpseudocode}
\usepackage{tikz}
\usepackage{float}
\usetikzlibrary{math,shapes}
\usepackage[caption=false, font=footnotesize, ]{subfig}
\usepackage{tabularx}
\usepackage{pbox}
\usepackage{ragged2e}
\usepackage{multirow}
\usepackage{tabularx}
\usepackage{balance}
\usepackage{subfig}
\usepackage{relsize}
\usepackage{pifont}
\usepackage{booktabs}
\usepackage{adjustbox}
\usepackage{cite}

\usepackage{booktabs} 
\usepackage{adjustbox} 

\usepackage{hyperref}

\usepackage{listings}
\mathchardef\mhyphen="2D

\usepackage{etoolbox}
\makeatletter
\patchcmd{\@makecaption}
  {\scshape}
  {}
  {}
  {}
\makeatletter
\patchcmd{\@makecaption}
  {\\}
  { -\ }
  {}
  {}
\makeatother
\usepackage{siunitx}

%% file: abstract.tex
Hexapod robots are potentially suitable for carrying out tasks in cluttered environments since they are stable, compact, and light weight. They also have multi-joint legs and variable height bodies that make them good candidates for tasks such as stairs climbing and squeezing under objects in a typical home environment or an attic. Expanding on our previous work on joist climbing in attics, we train a legged hexapod equipped with a depth camera and visual inertial odometry (VIO) to perform three tasks: climbing stairs, avoiding obstacles, and squeezing under obstacles such as a table. Our policies are trained with simulation data only and can be deployed on low-cost hardware not requiring real-time joint state feedback. We train our model in a teacher-student model with 2 phases: In phase 1, we use reinforcement learning with access to privileged information such as height maps and joint feedback. In phase 2, we use supervised learning to distill the model into one with access to only onboard observations, consisting of egocentric depth images and robot pose captured by a tracking VIO camera. By manipulating available privileged information, constructing simulation terrains, and refining reward functions during phase 1 training, we are able to train the robots with skills that are robust in non-ideal physical environments. We demonstrate successful sim-to-real transfer and achieve high success rates across all three tasks in physical experiments.

 

%% file: section-1-introduction.tex
\section{Introduction}
Lightweight legged robots are ideal platforms for navigating inside cluttered home environments in which the robot has to get around big objects such as a refrigerators, squeeze under low objects such as couches and beds, and climb over staircases to get from one floor to the other.  They can also be useful in rough environments such as attics, which are full of joists, and are uncomfortable and potentially dangerous for workers to do air sealing and vacuuming before they add insulation \cite{Zakhor_Lathrop_Austin_Zang_Tang_Roychowdry_Naing_Beauden_Li}. For example, since attics typically consist of multiple rows of joist structures, a human worker could easily fall through the attic floor and get seriously injured if they step on the sheetrock between two joists by mistake.
 
There has been a significant amount of work on
quadrupeds and bipedal robot locomotion. In our prior work \cite{10341957}, we developed methods to enable hexapods to climb joists in harsh attic environment. In this work, we extend our prior work to stair climbing, obstacle avoidance and squeezing under objects for a hexapod robot. We focus on hexapods for
two main reasons. First, hexapod robots can be more
stable and lightweight than quadrupeds and humanoids of
similar size. Secondly, bipedal or quadruped robots are often
taller than hexapods and are therefore
less suitable for traversing tight spaces such as the
corners of attics. To facilitate the practical usage of robots
in the retrofit business, it is important for legged locomotion
controllers to work with low-cost hardware. However, most
existing legged locomotion systems require high-end robots
capable of real-time sensing of joint states, which could
ultimately result in expensive hardware on the order of thousands if not tens of thousands of dollars. For example, model 
predictive control methods such as \cite{tan2018simtoreal} require powerful
computation resources and real-time joint feedback from
expensive robot platforms and often compromise real-time
performance when incorporating more complex dynamics.
Data-driven methods such as \cite{agarwal2022legged} can work with limited computation
resources and are robust to a variety of perception
failures but need fast joint state feedback. Many low-cost
robots are not equipped with powerful onboard computation
or real-time feedback such as joint torque and angle that
are accessible on more expensive platforms. Meanwhile,
humans without leg sensing feedback when equipped with
prosthetics can walk and even participate in competitive
sports with only egocentric visual perception and a sense
of body orientation \cite{Paramaguru_2012}. 
\input{figures/new_robot}
In this paper, we propose an
end-to-end learning-based perceptive controller for low-cost,
sub-thousand-dollar hexapods to autonomously climb over staircases, avoid obstacles and squeeze under objects and demonstrate zero-shot sim-to-real transfer in real environments. In addition to attics, these skill sets are useful for robots to navigate inside homes full of furniture which the robot has to get around, and squeeze under obstacles and climb staircases to get from one floor to the other. Our robot is a \$600 SpiderPi robot with no real time joint feedback manufactured by
Hiwonder, shown in Figure \ref{fig:robot}, equipped with an Intel L515
depth camera and a T265 tracking camera with a customized
camera mount. Similar to the approach proposed in \cite{10341957}, we use a two-stage teacher-student training procedure to learn models that can work without real-time
joint feedback: the first stage involves Reinforcement
Learning (RL) with access to privileged observations and
the second stage uses supervised learning to distill the model
using only onboard observations including body pose
and egocentric depth images. Since optimal stair climbing, squeezing and obstacle avoidance
motions are fundamentally different from walking, we train
our controllers without human-defined prior gait knowledge,
guiding the models to explore task-appropriate motions.
Through extensive simulations and physical experiments we show our low cost robot successfully learn the following three skills and generalize across different terrains:  (a) climb up and down staircases with as many as 15 steps including a landing pad ; (b) squeeze under objects that are low, regardless of whether the objects are long or short; (c) maneuver right and left around multiple obstacles without scarping or touching them and continue walking in the same direction it was headed before each obstacle. With
proper design of terrains, reward functions, and choice of privileged information in simulation environment, our model is able to learn a variety of control skills.

%% file: figures/new_robot.tex
\begin{figure}
    \vspace{0.2cm}
    \centering
\includegraphics[width=88mm]{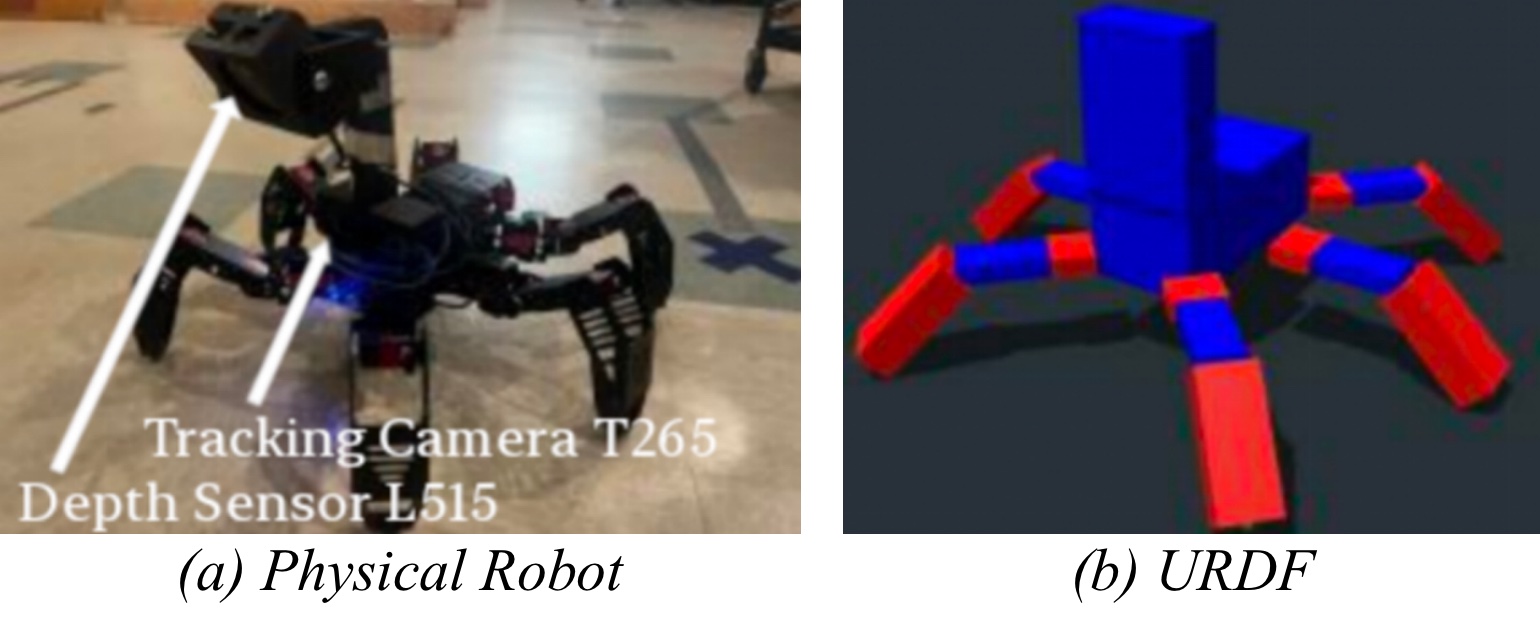}
    \caption{Physical and URDF of the robot. \textit{(a)} The hexapod robot standing at the reset position - roughly 37 cm tall. \textit{(b)} The URDF of the hexapod.}
    \label{fig:robot}
    \vspace{-0.5cm}
\end{figure}

%% file: section-2-related-work.tex
\section{Related Work}
In this part, we focus on relevant prior works in two major areas: (1) locomotion control for hexapods, quadrupeds, and robots of other shapes; (2) reinforcement learning for stair climbing, obstacle avoidance, and squeezing.
\subsection{locomotion control}
Prior works have focused on specific control methods to achieve basic locomotion abilities in hexapod and quadruped robots. Researchers have used two-layer Central Pattern Generator (CPG) networks and posture control strategies based on Force Distribution and Compensation to enable robot locomotion across a variety of terrains 
\cite{ouyang_adaptive_2021}. In our previous work, we successfully enabled the hexapod to climb over joist terrains using a teacher-student model and actor/critic reinforcement learning strategy \cite{10341957}.

Deep Reinforcement Learning (DRL) has become a promising approach for developing autonomous and complex behaviors in real world systems. Many researchers choose to test the performance of their systems in simulation environments before applying them to real-world applications. However, the sim-to-real gap poses significant challenges in transferring simulated learning to real-world applications \cite{ibarz_how_2021, rizzardo_sim--real_2023}. 
Solutions such as system identification, domain randomization, domain adaptation, imitation learning, meta-learning, transfer learning, and knowledge distillation have shown promise in narrowing this gap, enabling more effective deployment of robotic systems in physical environments\cite{ibarz_how_2021,hua_learning_2021,tan_sim--real_2018,zhao_sim--real_2020}.
Rizzardo proposed a sim-to-real technique that trains a Soft-Actor Critic agent together with a decoupled feature extractor and a latent-space dynamics model, enabling transferring without retraining or fine-tuning \cite{rizzardo_sim--real_2023}.  
Tiboni introduced DROPO, a novel method for estimating domain randomization distributions for safe sim-to-real transfer \cite{tiboni_dropo_2023}. 

Model-free reinforcement learning has emerged as a pivotal approach for achieving different locomotion tasks, such as hopping and crawling, in various environments, both terrestrial and aquatic, barriers or gaps \cite{song_deep_2021, shi_deep_2020, bogdanovic_model-free_2022, wan_learning_2023, zhuang_robot_2023, kareer_vinl_2023, yu_visual-locomotion_nodate}. 
Special network designs have been introduced to facilitate the training of the locomotion policy. 
\cite{wan_learning_2023} proposed learning low-level motion from a biological dog first and then learning high-level tasks in order to save training time. All these policies were combined into a single framework, allowing the robot to autonomously select and execute the appropriate policy.
\cite{kareer_vinl_2023} used two-layer structure, a visual navigation layer to output the angular velocity commands and a visual locomotion layer to control quadruped robots to  step over scattered ground to the destination. 
\cite{yu_visual-locomotion_nodate} proposed a hierarchical structure with a high-level vision policy and a low-level motion controller to enable a quadrupedal robot to traverse uneven environments.

\subsection{Reinforcement learning}
In order to show the efficacy of the policy employed on the robot, physical experiments are imperative. Traditional methods such as dynamic window approach (DWA)\cite{580977} are effective in avoiding obstacles robotics. Actor-critic reinforcement learning-based avoidance methods have been used to enable robots to avoid scattered obstacles \cite{choi_reinforcement_2021,hart_enhanced_2024}. \cite{Lee_2020} proposed training ANYmal robots with reinforcement learning in simulation and deployed the policy to run in challenging natural environments. \cite{kumar2021rma} presents an approach to teach quadruped robot to adapt and conquer unseen environments with a base policy and adaption module. Researchers introduced a general DRL framework for obstacle avoidance, and a manipulability index into the reward function in order to avoid joint singularity while executing tasks \cite{shen_reinforcement_2022}. The FAM-HGNN framework, which relies on an attention mechanism within a heterogeneous graph neural network, presents a novel solution for the obstacle avoidance problem in RL. This approach surpasses the performance of both multi-layer perceptron-based and existing GNN-based RL methods \cite{zhang_obstacle_2024}.

With regards to squeezing, researchers enabled the reconfigurable robot RSTAR to squeeze through two adjacent obstacles, duck underneath an obstacle and climb over an obstacle using Q learning algorithm \cite{yehezkel_overcoming_2020}.

In tasks involving climbing stairs, both Deep Deterministic Policy Gradients (DDPG) and Trust Region Policy Optimization (TRPO) were evaluated, with the latter showing superior performance \cite{garg_virtual_2018}. Researchers used sim-to-real RL to achieve climbing by only modifying an existing flat-terrain training framework to include stair-like terrain randomization, without any changes in reward function \cite{siekmann_blind_2021}. Researchers also performed experiments on enabling different articulated, tracked robots \cite{mitriakov_reinforcement_2021} and assistive robots \cite{mitriakov_staircase_2020} to climb on slopes or stairs using machine Learning techniques.

\input{figures/reward_terms}

%% file: figures/reward_terms.tex
\begin{table*}
\vspace{0.2cm}
\centering
\resizebox{\textwidth}{!}{%
\begin{tabular}{|l|l|l|l|l|l|}
\hline
Reward Term & Expression & Joist Climbing & Stair Climbing & Obstacle Avoidance & Obstacle Squeezing \\
\hline
Linear velocity in global x (forward) & $\text{clip}(v_x, \text{min} = -0.4, \text{max} = 0.4)$ & $1^2$ & $1^2$ & $1^2$ & $1^2$\\
Linear velocity in body y (left/right) & $v_y^2$ & $-1^1$ & $-1^1$ & $-1^1$ & $-1^1$\\
Global heading & $\theta^2$ & $-3^1$ & $0$ & $0$ & $0$\\
Angular velocity: yaw & $\omega^2$ & $-1^0$ & $0$ & $-1^0$ & $-1^0$\\
Ground impact & $\|f_t - f_{t-1}\|^2$ & $-1^{-1}$ & $0$ & $0$ & $0$\\
Collision penalty & $\mathbf{1}\{\text{coxa, femur, or base contacting terrain}\}$ & $-1^0$ & $0$ & $-3^0$ & $-1^0$\\
Action rate & $\|a_t - a_{t-1}\|^2$ & $-5^{-1}$ & $-5^{-1}$ & $-5^{-1}$ & $-5^{-1}$\\
Action magnitude & $\|a_t\|^2$ & $-1^{-2}$ & $0$ & $-1^{-2}$ & $-1^{-2}$\\
Torques & $\|\tau\|^2$ & $-1^{-3}$ & $0$ & $-1^{-3}$ & $-1^{-2}$\\
Joint acceleration & $\hat{\ddot{q}}^2 = \frac{\dot{q}_t - \dot{q}_{t-1}}{\Delta t}^2$ & $-1^{-5}$ & $-1^{-5}$ & $-1^{-5}$ & $-1^{-5}$\\
Joint limit penalty & $\text{clip}(q_t - q_{\text{min}}, \text{max} = 0) + \text{clip}(q_t - q_{\text{max}}, \text{min} = 0)$ & $-1^0$ & $-1^0$ & $-1^0$ & $-1^0$\\
End effector height & $\|z_{\text{end\_effector}}\|$ & $-1^{-1}$ & $0$ & $0$ & $0$\\
Global y deviation & $\|y_{\text{current}} - y_{\text{start}}\|^2$ & $0$ & $-1^2$ & $-1^0$ & $-1^0$\\
Distance to obstacle (front) & $f(H)$ & $0$ & $0$ & $-1^{-1}$ & $0$\\
Distance to obstacle (above) & $f(H)$ & $0$ & $0$ & $0$ & $-1^{-1}$\\
\hline
\end{tabular}
}
\caption{Reward function terms and their corresponding weights for different tasks. The definition of the top 12 reward terms are in \cite{10341957}.}
\label{table:reward_terms}
\end{table*}

%% file: section-4-methodology.tex
\section{Methodology}
We aim to train the robot to climb up/down stairs, avoid obstacles, and squeeze under objects. For each task, we train the corresponding policy in simulation using Isaac Gym, then directly deploy the trained policy onto the robot, which is equipped with an Intel RealSense Tracking Camera T265 for pose estimation and an L515 for depth estimation. Each task requires a different terrain construction, reward function, and camera angle, as shown in Tables \ref{table:reward_terms} and \ref{table:camera}. We adopt a 2-stage student-teacher model shown in Figure \ref{fig:diagram}, similar to \cite{10341957} in which the teacher policy is trained with privileged information such as joint feedback and height map shown as the yellow pad in Figure \ref{fig:sim}, and is transferred to a student policy that takes in pose estimation and depth images as visual input using supervised learning. In the teacher model training, even though different tasks are trained with different reward terms, sizes of height map, and terrain constructions, they are each distilled to a policy that takes depth map of size $320 \times 240$ as visual input. During training, we control our robot by directly predicting the joint angles of the robot and applying them to the corresponding joints. The angles range from [-120, 120] degrees. Each joint is initialized to a resting position provided by the manufacturer as shown in Figure \ref{fig:robot}.
\input{figures/diagram}
\input{figures/simulation}
The entire training process takes approximately 10 hours for the teacher policy and 5 hours for the student policy on an RTX TITAN GPU. The reward terms and weights for each of the four tasks, joist climbing, stairs climbing, obstacle avoidance, and squeezing, are shown in Table \ref{table:reward_terms}. Different tasks require different number of training iterations to converge, as seen in the reward convergence curve shown in Figure \ref{fig:reward}. As seen, the squeezing task converges in 10 times fewer episodes than the other two tasks. During model deployment, we take the depth image and pose estimation from our visual odometry system as input to our policy and output the joint angles for each of the 18 joints. We then send signals to each servo to set them to desired angles. 
\input{figures/reward}
\input{figures/camera}

The optimal camera angle for different tasks is shown in Table \ref{table:camera}. Other than squeezing, which requires the optical axis of the camera to be parallel to the horizon, all other tasks work well with a 30-degree downward-looking inclination. This is to be expected since the robot has to look “up” rather than “down” to see the obstacle above it, when it is squeezing under an object. 

Height map size and location for different tasks are also shown in Table II. The latter refers to the distance of height map from the front of the robot. We have empirically found obstacle avoidance to need a larger height map to be further away from the robot than joist climbing and stair climbing.

\subsection{Stair climbing}
We trained the robot to climb up and down staircases according to the 2-step teacher-student method described earlier. We deployed curriculum training in order to help the robot climb more  challenging staircases by first learning easier ones.  In particular, we let the riser heights increase from 4.5 cm to 18 cm and the tread depth decrease from 30 cm to 18 cm as the level of difficulty increases in curriculum training. In addition,  we randomized tread depth in a given staircase in simulation to improve generalization of our policy. 

The  weight of the reward term in the third row from the bottom in Table \ref{table:reward_terms} is an order of magnitude larger for stair climbing than the other tasks. The main motivation for this is to keep the robot from unnecessarily deviating to the right or left as it climbs up a staircase. Intuitively this term minimizes the lateral deviation of the robot from the direction of the axis it was pointed before it starts climbing. We empirically found that the 30° tilt angle for the  depth camera works well for both climbing up and down staircases. 

\subsection{Obstacle avoidance}
Even though traditional methods such as DWA\cite{580977} work well in obstacle avoidance, since our eventual goal is to combine the different RL  skills into one, we will also develop an RL based policy for 
obstacle avoidance.
The most straightforward way to teach the robot to avoid collisions is by minimizing collisions between the robot’s femur, coxa, and body during simulation, assigning large negative rewards when such collisions occur. This is shown as the “Collision Penalty” in Table \ref{table:reward_terms} and indicates the number of collisions, whereby a collision is defined as an event with contact force larger than a certain force threshold. We have empirically found that such a method results in the robot scraping by and touching  obstacles as it tries to avoid them.  To circumvent that,  we use a reward term shown in the one to the last row of Table \ref{table:reward_terms}  given by: 
\vspace{-1pt}
$$
R(H) = - \sum_{i=1}^{M} \left( \sum_{j=1}^{N} h'_{ij} \cdot \overrightarrow{w_{1}}_j \right) \cdot \overrightarrow{w_{2}}_i
\vspace{-1pt}
$$

where $h'_{i,j}$ is the $(i,j)^{th}$ element of  the binarized height map $M \times N$ matrix $H'$ given by:
$$
h'_{ij} = 
\begin{cases} 
1 & \text{if } h_{ij} > 0 \\
0 & \text{otherwise}
\end{cases}
\vspace{-3pt}
$$
$h_{ij}$ is the  $(i,j)^{th}$ element of the $M \times N$ matrix height map matrix $H$, and $\overrightarrow{w_1}$ is an $N$ dimensional weight vector of dimension $N$ of a triangular shape given by: 
$$\overrightarrow{w_1} = \left[1, 1 + \frac{2}{N}, 1+\frac{4}{N}, \ldots, 2, \ldots, 1 + \frac{4}{N}, 1 + \frac{2}{N}, 1\right]$$
corresponding to  the perpendicular distance of the points in the binarized height map to the axis parallel to the direction of the movement of the robot passing through its center.
$w_2$ is an $M$ dimensional weight vector of the shape of a ramp ranging from 1 to 2, providing more weight to the obstacle points closer to the front of the robot given by:
$$\overrightarrow{w_2} = \left[1, 1 + \frac{1}{M}, 1 + \frac{2}{M}, \ldots, 2\right]
$$

\input{figures/squeezing_design}
For the terrain in simulation, we included a variety of obstacle shapes in the map. We also note that if the obstacles are densely placed on the map at the beginning stages of the training, the policy likely converges to a local maximum reward and ends early. This is because of the sudden large penalty generated from distance to obstacle reward term. Thus, we incorporate curriculum training where the robot first learns to walk without any obstacles and then to avoid sparsely placed obstacles followed by a more dense obstacle terrain. The density ranges of each level of difficulty is given by $(2 \times Level/Total\_Levels) \times Density_{final}$, where the $Density_{final}$ is defined as the density of the obstacle spacing in the final difficulty level.

The obstacle avoidance functionality provided by the manufacturer of our robot uses  ultrasound to sense obstacles and change direction accordingly, but it never goes back to its original orientation as seen in this video \href{https://www.youtube.com/shorts/irWzEnhbGC0}{link}. In contrast, we ideally want our obstacle avoidance functionality to have the robot revert back to its original orientation after it avoids a given obstacle. To achieve this, we use a combination of global y deviation term and a local y velocity term where the y axis is perpendicular to the direction the robot is headed and moving along before it reacts to an obstacle. Global y deviation term, shown in third to the last row of Table \ref{table:reward_terms}, indicates the distance between the robot and the y axis. The local y velocity term shown in row 2 refers to the square velocity along the y direction. The latter term prevents the robot from going back to the original direction of motion too soon due to not having the obstacle in its field of view after the initial turn. 


\subsection{Squeezing under obstacles}
Squeezing under objects is an important skill to learn in traversing environments with low objects such as beds, couches, and tables. We mainly care about two aspects in the design of the squeezing terrain; obstacle height and ’tunnel length’ as defined in Figure \ref{fig:squeeze}. We aim to make the robot learn such a skill and generalize to obstacles of varied shapes and ’tunnels’ with varied lengths.

Since Isaac Gym does not support floating terrains and requires everything  to be grounded, we modified the code to construct terrains on vertices 'in the air'. In the simulation environment, we created a floating obstacle above the basic terrain shown in Figure \ref{fig:sim}(c) to mimic the effect of a physical table, bed, or couch. The height map used in phase 1 of the training in simulation, takes on the ground level height when there is no obstacle above and the obstacle’s height otherwise. We then define a single reward term, shown in the last row of Table \ref{table:reward_terms}, to maximize the distance between the robot’s body to the obstacle when there’s an obstacle above and maximize the distance between the robot’s body to the ground otherwise. Concretely, the reward is defined as follows:
\vspace{-5pt}
$$
R(H) = - \sum_{i=1}^{N} \left( \sum_{j=1}^{M} h''_{ij} \cdot \overrightarrow{w_{3}}_j \right) \cdot \overrightarrow{w_{4}}_i
$$
where $h''_{ij}$ is a function of the height map $h_{ij}$ and the distance of the base robot to the ground, $b$, as follows:

$$
h''_{ij} = 
\begin{cases} 
1 & \text{if } (h_{ij} - b) > 0 \\
-2 \cdot |h_{ij} - b| & \text{otherwise}
\end{cases}
$$
$\overrightarrow{w_3}$ is a $N$ dimensional vector of all ones, and $\overrightarrow{w_4}$ is an $M$ dimensional ramp vector ranging from 1 to 2 to assign a larger weight to the obstacle points closer to the front of the robot given by:
\vspace{-3pt}
$$\overrightarrow{w_4} = \left[1, 1 + \frac{1}{N}, 1 + \frac{2}{N}, \ldots, 2\right]$$

We use curriculum learning so that the robot first learns to walk and then squeeze under the obstacles. The level of difficulty increases as we decrease the object distance to the ground from 37 to 35 to 33 and to 31 cm. The obstacle itself can have variable height in the vertical direction as seen in Figures \ref{fig:squeeze_s}(a) and \ref{fig:squeeze_s}(b), and variable length or width. The longer the obstacle is, the longer the robot has to squeeze to avoid hitting it while it squeezes underneath. These three parameters are  randomly chosen across all levels since they do not reflect the difficulty of the task. We also randomize the origin so that in simulation the robot reaches the obstacle after walking different distances. Finally, the optical axis of the depth camera is parallel to the ground so that it can clearly see the beginning and the end of the “tunnel” created by the obstacle. 


%% file: figures/diagram.tex
\begin{figure}
    \vspace{0.2cm}
    \centering
    \includegraphics[width=88mm, clip]{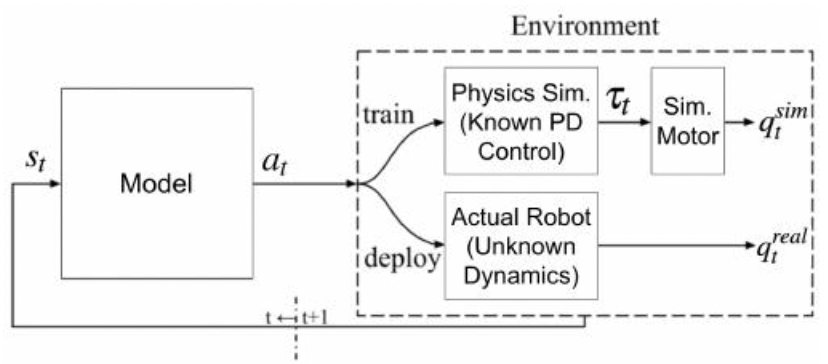}
    \caption{ High-level overview of training methodology \cite{10341957}}
    \label{fig:diagram}
    \vspace{-0.3cm}
\end{figure}

%% file: figures/simulation.tex
\begin{figure}
    \vspace{0.2cm}
    \centering
    \includegraphics[width=84mm, clip]{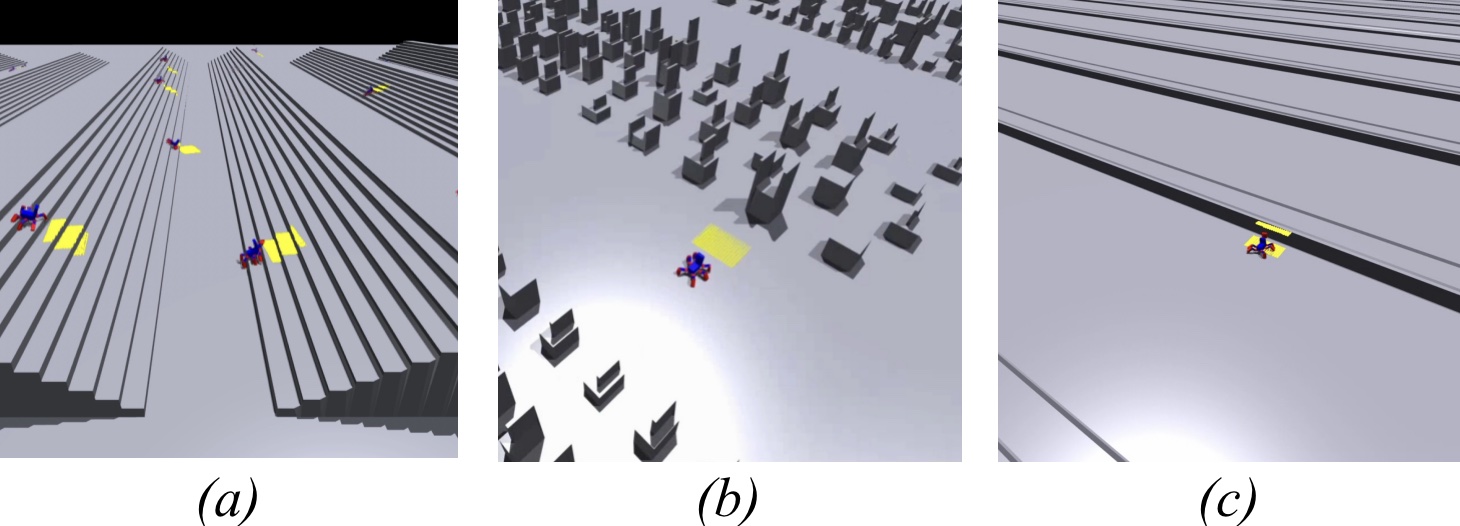}
    \caption{Simulation environment in Isaac Gym. \textit{(a)}  Stairs Climbing. \textit{(b)}  Obstacle Avoidance. \textit{(c)}  Squeezing under obstacles.}
    \label{fig:sim}
    \vspace{-0.5cm}
\end{figure}

%% file: figures/reward.tex
\begin{figure}
    \vspace{0.2cm}
    \centering
    \includegraphics[width=84mm, clip]{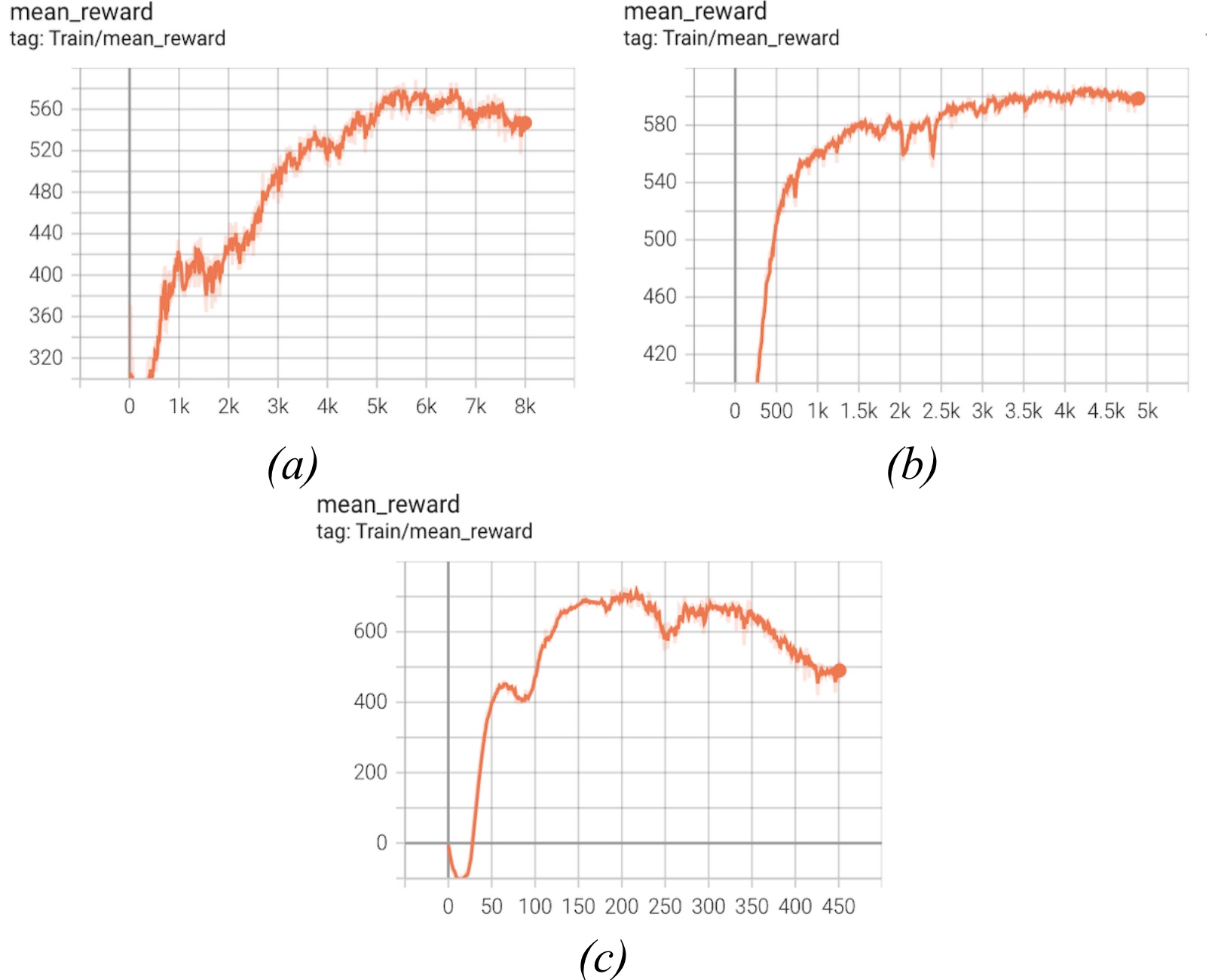}
    \caption{Reward vs. episode convergence curve for tasks. \textit{(a)} Stairs Climbing. \textit{(b)} Obstacle avoidance. \textit{(c)} Squeezing under obstacles.}
    \label{fig:reward}
    \vspace{-0.5cm}
\end{figure}

%% file: figures/camera.tex
\begin{table*}[h]
\vspace{0.2cm}
\vspace{+3pt} 
\centering
\begin{adjustbox}{width=0.8\textwidth} 
\begin{tabular}{@{}l|c|c|c@{}} 
\toprule
\textbf{Task} & \textbf{Camera Angle (degrees)} & \textbf{Height map size (m)} & \textbf{Height map location (m)} \\ 
\midrule
Joist Climbing & 30 & $0.6 \times 0.8$ & 0.3 \\
Stairs Climbing & 30 & $0.6 \times 0.8$ & 0.3 \\
Obstacle Avoidance & 30 & $0.6 \times 1.0$ & 0.6 \\
Squeezing under Obstacles & 0 & $0.6 \times 0.8$ & 0 \\
\bottomrule
\end{tabular}
\end{adjustbox}
\caption{Optimal camera angle and height map for each task.}
\label{table:camera}
\vspace{-5pt} 
\end{table*}

%% file: figures/squeezing_design.tex
\begin{figure}
    \vspace{0.5cm}
    \centering
    \includegraphics[width=46mm, clip]{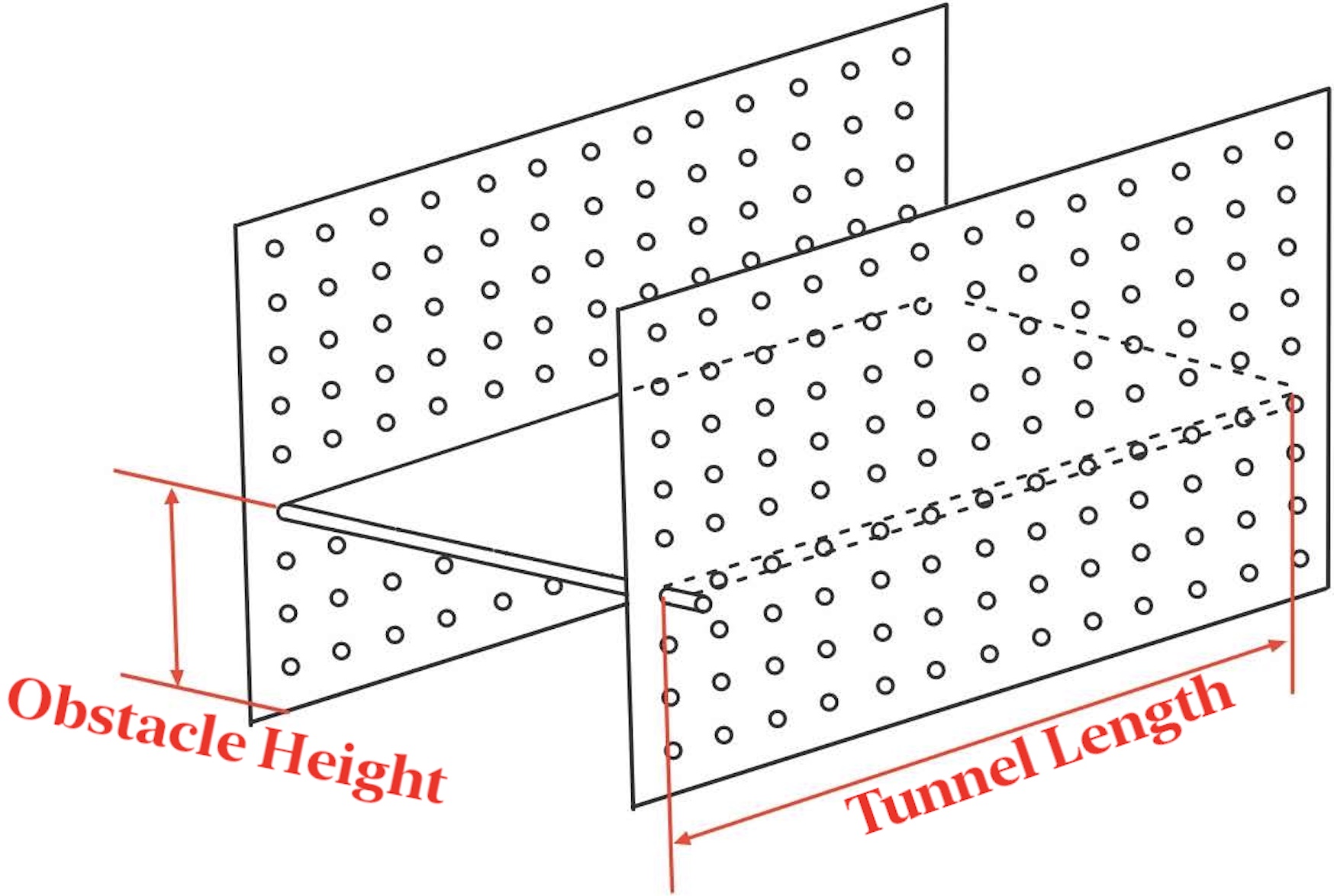}
    \caption{Physical design of a squeezing environment.}
    \label{fig:squeeze}
    \vspace{-0.5cm}
\end{figure}

%% file: section-5-experiments.tex
\newcommand{\cmark}{\ding{51}} 
\newcommand{\xmark}{\ding{55}} 

\input{figures/stairs_table}
\input{figures/obstacles_table}
\section{Physical Experiments}
We run the policies trained in simulation on a Raspberry Pi processor placed on the physical robot. We conducted experiments on all three tasks to verify and characterize the performance of our policies in real-world environments. For the staircase, we used three sets on the campus of U.C. Berkeley and for the other two tasks we constructed  several terrains to evaluate the performance of the robot by measuring its success rate in completing tasks.

\subsection{Stair climbing}
\input{figures/stairs_scene}

We chose three staircases on the U.C. Berkeley campus, as shown in Figure \ref{fig:stairs_s}, to evaluate the robot's performance in both climbing up and descending tasks. In Figure 5, Cory Hall and Soda Hall each had 7 steps and Sutardjai Hall had 8 steps. At the start of each experiment, the robot is placed 20 cm in the front of the stairs and is reset to its default standing position shown in Figure \ref{fig:robot}. We interrupt only if the robot falls over or is stuck on a stair.

We evaluated performance by counting the number of steps the robot could complete in each trial. We then average that number across 10 trials as shown in Table \ref{table:climbing_table}. As seen, Soda Hall has the highest success rate because it has the lowest rise and the highest tread. Few of the climbing up failure cases resulted from the robot falling over. Others resulted from the robot getting stuck on the last stair for too long, where it believes it is walking on a plane ground as the camera image is not showing any stairs. For climbing down stairs, the failure cases come from the robot losing control and falling down. We present a video \href{https://youtu.be/Z22-K-4Pe6E}{link} showing the robot climbing two sets of stairs separated by a platform, with 8 and 7 steps in the first and second sets respectively. As seen in the video, the robot is able to traverse in a relatively straight line in the center without too much lateral shift to right or left. The video also shows the robot climbing down stairs with our policy compared to a baseline walking policy where it is trained to walk on a flat terrain environment. As expected, applying the "walk" policy to the staircase results in the robot crashing down the stairs.
\input{figures/obs_scene}
\subsection{Avoiding obstacles}
We constructed a variety of obstacle scenarios to evaluate the robot’s obstacle avoidance performance. We test the robot’s generalization ability to detect arbitrary shapes of obstacles and to avoid robustly. We design a total of four environments. The first one consists of a single box-shaped obstacle placed in front of the robot. The second consists of two obstacles: a box-shaped obstacle placed in front and a deformable plastic bag placed behind it. The third consists of a person standing/sitting in front of the robot. The fourth consists of a person walking towards the robot. We test each of the environments 10 times, with our results in Table \ref{table:o_1}. We group the outcome into three cases: success, scrape, and collision. Success means the robot can navigate around the obstacle without any collision, and scrape means the robot is able to go around the obstacle but with some leg scraping the objects/person, and collide means the robot runs into the object. For the failure cases, the robot often detects the objects, but reacts too slowly with the back leg scraping the objects after the front part successfully avoids it. In success cases, the robot exhibits avoidance behavior both to the left and right depending on its distance to each side of the obstacle. This indicates our policy did not merely memorize to avoid objects by always veering to one direction. After passing the obstacle, the robot rotates back to its original orientation and keeps moving forward. Finally, we did a comprehensive test on the robot’s ability to avoid all six of our tested’s obstacles in a row as shown in Figure \ref{fig:obs_s} and visualized in this video \href{https://youtu.be/tbSoOhQQsJk}{link}. 

\input{figures/squeeze_scene}
\subsection{Squeezing under obstacles}
The squeezing setup shown in Figure \ref{fig:squeeze_s}, aims to mimic obstacles such as a couch, a bed, or a table. At the beginning of the experiment, the robot is positioned 30 cm in front of the obstacle. It is tasked to squeeze under the object and walk in the squeezed mode until it gets out of the “tunnel” created by the overhead obstacle at which time it is expected to raise its body to the default height and continue walking. The experiment is considered a success if the robot is able to cross the tunnel without any collisions\footnote{Collisions are visually detected and defined as events where contact between the robot and obstacles results in a trajectory change.}. Our experimental set up for squeezing is shown in Figure \ref{fig:squeeze_s} with the dimensions superimposed on the pictures. In Figures \ref{fig:squeeze_s}(a), \ref{fig:squeeze_s}(c) and \ref{fig:squeeze_s}(d) there are a pair of metal rods that are 4" or 10.20 cm apart. The end of the rods are plugged into two boards with holes on a one inch grid. In Figures \ref{fig:squeeze_s}(c) and \ref{fig:squeeze_s}(d) there are paper pieces connecting the two rods, creating a different depth image from Figure \ref{fig:squeeze_s}(a) during inference and hence testing the generalizability of our policy. Finally, there is a wood board in Figures \ref{fig:squeeze_s}(b) and \ref{fig:squeeze_s}(d) creating a tunnel of length 129.1 cm to ensure the robot can be in the squeeze position for an extended time. As a reference, the robot is 28.75 cm high when laying flat and 37 cm high when standing in the reset standing position shown in Figure \ref{fig:robot}.  While we acknowledge that the height of the robot in the squeezing position is still too large, our main goal is to demonstrate the concept of teaching the robot to squeeze when needed.
\input{figures/s_s}For each of the four settings shown in Figure \ref{fig:squeeze_s}, we repeat 10 to 20 trials and present the success rate in Table \ref{table:s1}. \input{figures/squeezing_table}

As seen, the success rate is high\footnote{We have empirically found that the failure rate for all three tasks increases when the battery is low and the supplied voltage is around 10 rather than the nominal 12 volts.}. While the robot can instantaneously squeeze as low as 31.75 cm to go under a thin metal rod, it cannot sustain that height for an extended distance of say 129 cm without hitting the tunnel roof. However, it can sustain the squeeze posture for a 34.29 cm tunnel of Figure \ref{fig:squeeze_s}(b). The support boards on the side have grid hole spacing of an inch; as such we can only move the rods one inch at a time in the vertical direction, resulting in a "jump" from 31.75 to 34.29 cm height. We speculate that the robot can successfully go through a tunnel of height 33 cm without scraping the top.

In one experiment as shown in Figure \ref{fig:s_env} where there are two sets of obstacles of height 31.75 cm and no tunnel in between, we notice that the robot squeezes under the first obstacle, rises up shortly during the middle, squeezes again under the second obstacle upon sensing it, and finally stands up to walk away. We present a video demonstration of this experiment in this \href{https://youtu.be/dNtgt4t8Pa8}{link}. As seen in the video, the "belly" of the robot comes close to the ground as it squeezes under the obstacles and raises up afterwards.\input{figures/height_plot}
The plot in Figure \ref{fig:height_plot2} shows the change in the height of the robot's base relative to the ground, as captured in simulation.

%% file: figures/stairs_table.tex
\begin{table*}[t] 
\vspace{0.2cm}
\centering 
\begin{tabular}{@{}l|ccc|ccc@{}}
\toprule
\multicolumn{1}{c}{} & \multicolumn{3}{c}{Climbing Up (Stairs completed/total stairs)} & \multicolumn{3}{c}{Climbing Down (Stairs completed/total stairs)} \\
\cmidrule(lr){2-4} \cmidrule(lr){5-7}
Method & Cory Stairs & Soda Stairs & Sudardja & Cory Stairs & Soda Stairs & Sudardja \\
\midrule
Perceptive (Ours) & 6.0/7.0 & 7.0/7.0 & 7.6/8.0 & 6.6/7.0 & 7.0/7.0 & 7.5/8.0 \\
\bottomrule
\end{tabular}
\caption{Stair climbing performance.}
\label{table:climbing_table}
\end{table*}

%% file: figures/obstacles_table.tex
\begin{table*}[t]
\centering

\begin{tabular}{@{}l|ccc|ccc|ccc|ccc@{}}
\toprule
& \multicolumn{3}{c|}{Single Box} & \multicolumn{3}{c|}{A Box and A Bag} & \multicolumn{3}{c|}{Single Person Standing} & \multicolumn{3}{c}{Single Person Moving} \\
\cmidrule(lr){2-4} \cmidrule(lr){5-7} \cmidrule(lr){8-10} \cmidrule(lr){11-13}
Method & success & scrape & collision & success & scrape & collision & success & scrape & collision & success & scrape & collision \\
\midrule
Perceptive (Ours) & 7/10 & 2/10 & 1/10 & 9/10 & 1/10 & 0/10 & 7/10 & 2/10 & 1/10 & 8/10 & 1/10 & 1/10 \\
\bottomrule
\end{tabular}

\caption{Object avoidance performance.}
\label{table:o_1}
\end{table*}

%% file: figures/stairs_scene.tex
\begin{figure}
    \vspace{0.2cm}
    \centering
    \includegraphics[width=84mm, clip]{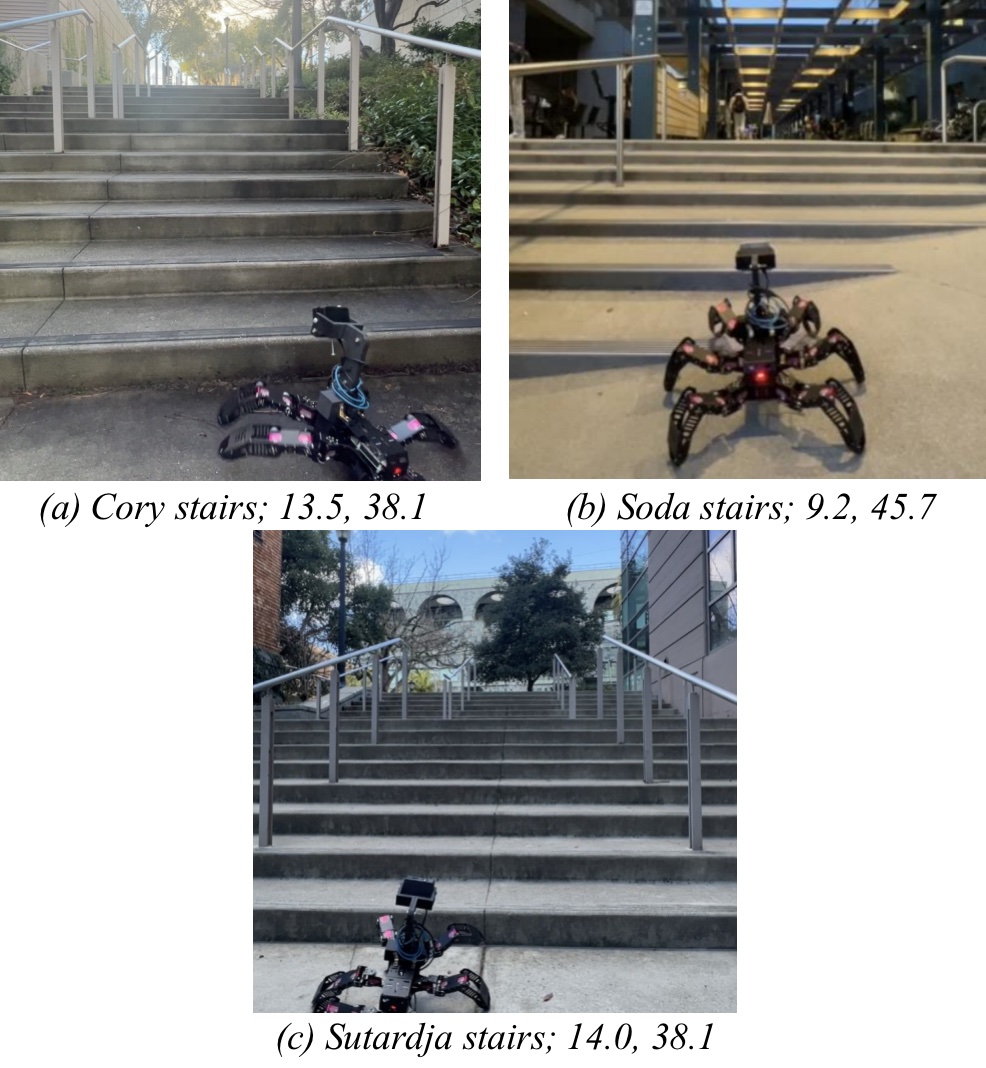}
    \caption{Experiments Scene for stairs climbing. The first number is the riser height and the second number is the tread depth both in centimeters.}
    \label{fig:stairs_s}
    \vspace{-0.5cm}
\end{figure}

%% file: figures/obs_scene.tex
=
\begin{figure}
    \vspace{0.2cm}
    \centering
    \includegraphics[width=40mm, clip]{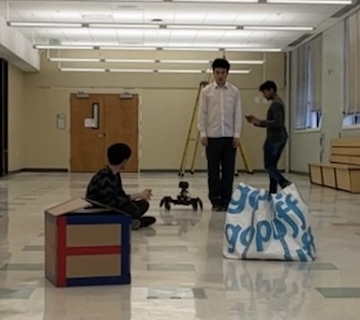}
   \caption{Obstacle Avoidance Environment.}
    \label{fig:obs_s}
    \vspace{-0.5cm}
\end{figure}

%% file: figures/squeeze_scene.tex
\begin{figure}
    \vspace{0.2cm}
    \centering
    \includegraphics[width=84mm, clip]{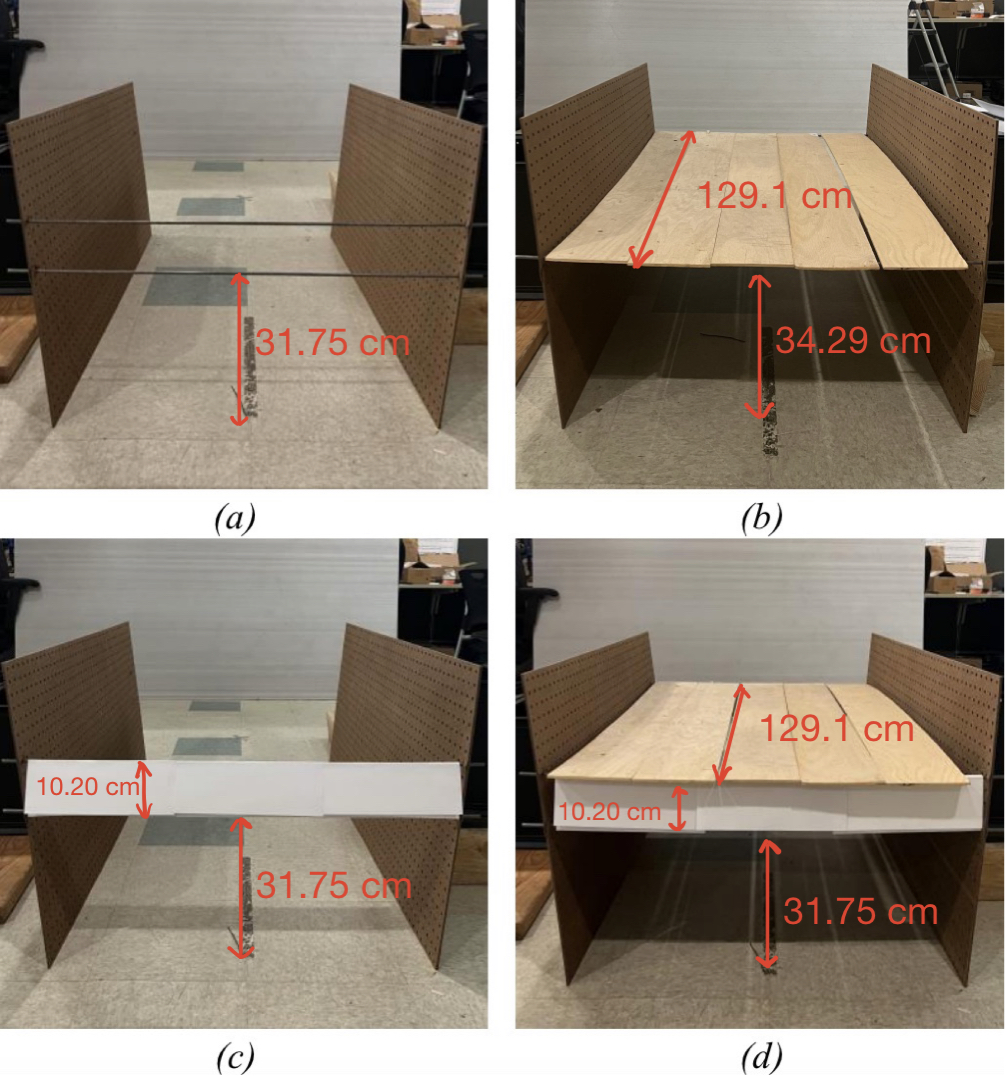}
    \caption{Experiments Scene for Squeezing. \textit{(a)} Metal rod as obstacle without tunnel. \textit{(b)} Lengthy tunnel. \textit{(c)} Paper block as obstacle without tunnel. \textit{(d)} Paper block as obstacle with tunnel.}
    \label{fig:squeeze_s}
    \vspace{-0.5cm}
\end{figure}

%% file: figures/s_s.tex
\begin{figure}
    \vspace{0.2cm}
    \centering
    \includegraphics[width=88mm, clip]{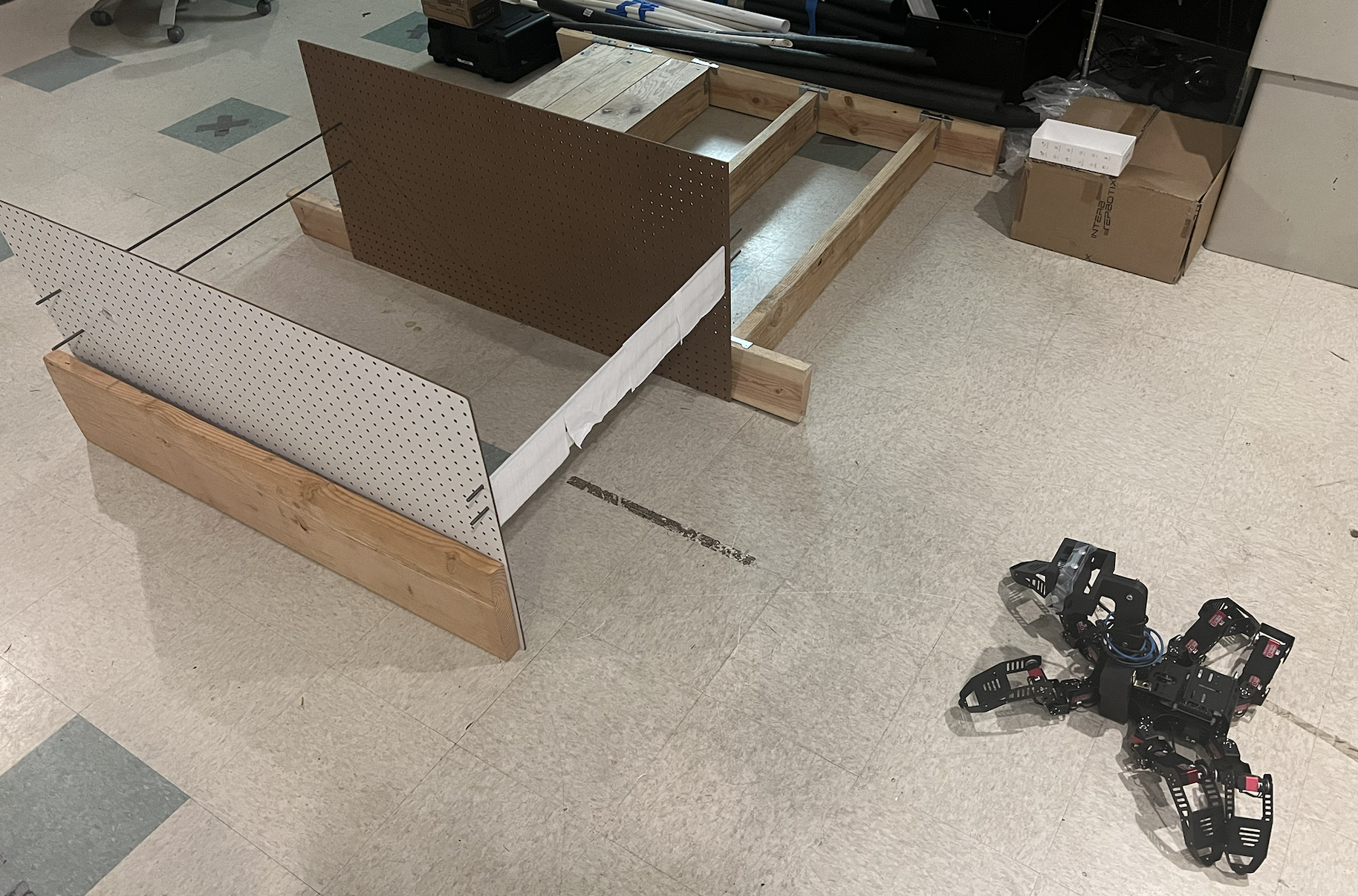}
    \caption{Squeezing under two consecutive obstacles.}
    \label{fig:s_env}
    \vspace{-0.5cm}
\end{figure}

%% file: figures/squeezing_table.tex
\begin{table}[ht]
\vspace{-3pt} 
\centering
\begin{adjustbox}{width=0.4\textwidth} 
\begin{tabular}{@{}l|c|c@{}} 
\toprule
\textbf{Terrain} & \textbf{Success Rate} & \textbf{\% Success} \\ 
\midrule
Metal rod w/o tunnel (a) & 17/20 & 85\% \\
Lengthy tunnel (b) & 10/10 & 100\% \\
Block w/o tunnel (c) & 18/20 & 90\% \\
Block with tunnel (d) & 9/10 & 90\% \\
\bottomrule
\end{tabular}
\end{adjustbox}
\caption{Squeezing performance with corresponding construction in Figure \ref{fig:squeeze_s}.}
\label{table:s1}
\vspace{-10pt} 
\end{table}

%% file: figures/height_plot.tex
\begin{figure}[ht]
    \centering
    \hfill
    \begin{minipage}{0.5\textwidth}
        \centering
        \includegraphics[width=84mm]{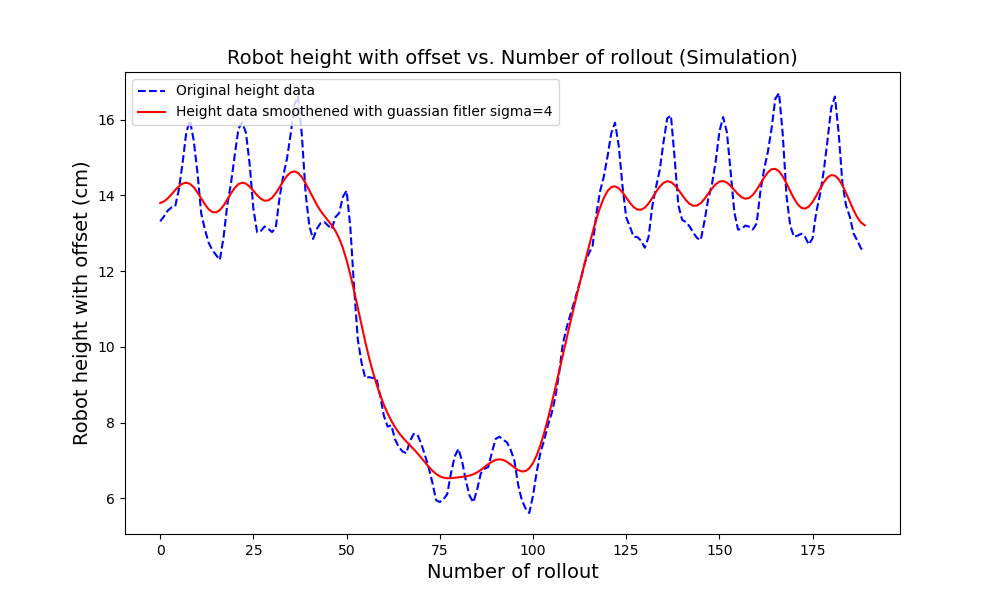} 
        \caption{Height of robot body vs. rollout step in Isaac Gym.}
        \label{fig:height_plot2}
    \end{minipage}
\end{figure}

%% file: section-7-limitations-and-future-work.tex
\section{Limitations}
One limitation of our approach is the requirement for different camera angles for climbing stairs and squeezing under objects. Switching policies on the go would thus present a bottleneck. Also, for squeezing under objects, although we have demonstrated the ability to generalize and to detect different shapes of objects and lengths of tunnels, the overall height of the system limits its performance to squeeze under extremely short obstacles such as a couch. The hexapod itself is quite short but the payload with the depth camera is too tall in the current system and limits the size of the objects it can squeeze under. In future work, we aim to incorporate an adjustable camera setup that extends upwards when the robot walks and downwards when it squeezes. We also need to change the camera angle between 0 and 30 degrees in the process. 

%% file: section-8-conclusion.tex
\section{Conclusions}
In this work, we trained a hexapod robot to climb up/down stairs, avoid obstacles, and squeeze under objects using only a depth camera and a visual inertial odometry sensor. We demonstrated the robot's ability to perform these tasks effectively through rigorous physical experiments. Future work consists of creating one universal policy combining all three tasks.